\documentclass[%
 reprint,
 superscriptaddress,
 amsmath,amssymb,
 aps,
floatfix,
]{revtex4-2}
\usepackage{amsfonts}
\usepackage{amsmath}
\usepackage{natbib}
\usepackage{subfigure}
\usepackage{float}
\usepackage{multirow}
\usepackage{graphicx}
\usepackage{dcolumn}
\usepackage{algorithmicx,algorithm}
\usepackage{array}
\usepackage{mathptmx}
\usepackage{algpseudocode,caption}

\usepackage{bm}
\begin{document}
\preprint{APS/123-QED}

\title{Autoregressive GNN-ODE GRU Model for Network Dynamics}

\author{Bo Liang}
\affiliation{Department of Automation, Shanghai Jiao Tong University, and Key Laboratory of System Control and Information Processing, Ministry of Education of China, Shanghai 200240, P. R. China}
\author{Lin Wang} 
\affiliation{Department of Automation, Shanghai Jiao Tong University, and Key Laboratory of System Control and Information Processing, Ministry of Education of China, Shanghai 200240, P. R. China}

\author{Xiaofan Wang}
\email{Corresponding author: xfwang@sjtu.edu.cn}
\affiliation{Department of Automation, Shanghai Jiao Tong University, and Key Laboratory of System Control and Information Processing, Ministry of Education of China, Shanghai 200240, P. R. China}
\affiliation{Shanghai Key Laboratory of Power Station Automation Technology, School of Mechatronic Engineering and Automation, Shanghai University, Shanghai, China}

\date{\today}

\begin{abstract}
Revealing the continuous dynamics on the networks is essential for understanding, predicting, and even controlling complex systems, but it is hard to learn and model the continuous network dynamics because of complex and unknown governing equations, high dimensions of complex systems, and unsatisfactory observations. Moreover, in real cases, observed time-series data are usually non-uniform and sparse, which also causes serious challenges. In this paper, we propose an Autoregressive GNN-ODE GRU Model (AGOG) to learn and capture the continuous network dynamics and realize predictions of node states at an arbitrary time in a data-driven manner. The GNN module is used to model complicated and nonlinear network dynamics. The hidden state of node states is specified by the ODE system, and the augmented ODE system is utilized to map the GNN into the continuous time domain. The hidden state is updated through GRUCell by observations. As prior knowledge, the true observations at the same timestamp are combined with the hidden states for the next prediction. We use the autoregressive model to make a one-step ahead prediction based on observation history. The prediction is achieved by solving an initial-value problem for ODE. To verify the performance of our model, we visualize the learned dynamics and test them in three tasks: interpolation reconstruction, extrapolation prediction, and regular sequences prediction. The results demonstrate that our model can capture the continuous dynamic process of complex systems accurately and make precise predictions of node states with minimal error. Our model can consistently outperform other baselines or achieve comparable performance.
\end{abstract}
\maketitle

\section{Introduction}
Complex networks, as the modeling of complex systems, are ubiquitous in real life, involving different disciplines and domains, such as physics~\cite{newman2003structure}, biology~\cite{liu2020computational}, sociology~\cite{kossinets2006empirical}, computer science~\cite{perozzi2014deepwalk} and control science~\cite{yuan2022mean} . Complex networks are essential carriers of information and have attracted much attention from researchers in recent years~\cite{benson2016higher}. Researchers attempt to utilize various tools to exploit and make good use of the information hidden behind the network. Besides, what interests researchers more is the dynamic process on the complex networks~\cite{gao2016universal}, that is, the evolution of the complex systems. Each unit in the network interacts with other units along the edges according to its specific dynamic rules, and the state of each unit evolves over time. Usually, the dynamic rules of each unit can be represented by the ordinary differential equation(ODE) or partial differential equation(PDE). For instance, in the gene regulatory network, the expression level of one gene is affected or restricted by the expression level of other genes, and the expression levels of genes change dynamically~\cite{alon2019introduction}. In the mutualistic interaction network, the number of a species is not merely affected by the ecological environment and external factors but also by other partners with mutualistic interactions~\cite{gao2016universal}. The fundamental and essential part of the dynamic process on the network is the governing equation which controls and models the dynamic laws of all the units. Discovering governing equations helps to understand the dynamics of the physical process, which also provides underlying guidance to the advancement of technology in return~\cite{jiahao2021knowledge}.

However, in many current dynamic systems on networks, governing equations are relatively unknown or partially known, which causes great difficulties in understanding and quantitative descriptions of the system. With the development of technology, states of the dynamic systems are measurable and traceable with instruments and sensors, and abundant time series data are observed, which gives rise to a new mode of data-driven discovery in governing equations or system identification~\cite{kaheman2020sindy}. Earlier researches include classic linear approaches~\cite{ljung2010perspectives}, symbolic regression~\cite{schmidt2009distilling}, dynamic modeling decomposition~\cite{kutz2016dynamic}, nonlinear regression~\cite{voss1999amplitude}, and nonlinear Laplacian spectral analysis~\cite{yair2017reconstruction}. Sparse regression also makes good advancement in this field~\cite{vlachas2018data}. Recent efforts are devoted to leveraging artificial neural network (ANN) and deep neural network(DNN) to tackle the time series data and identify the system model~\cite{vlachas2018data}. Champion et al.~\cite{champion2019data} present a deep auto-encoder neural network to model the coordinate transformation and utilize the sparse identification of nonlinear dynamics algorithm for parsimonious modeling. Nevertheless, these methods are susceptible to the system size and are not applicable to the complex dynamic process and large networks composed of thousands of units. In practice, node states of the network can not be observed ideally at regular intervals, and the time series data is usually irregular and sparse in most cases, which also causes problems for these methods. Nowadays, instead of identifying the dynamic model in mathematical functions, researchers try to simulate, fit and learn the network dynamics in the form of a black box without need of identifying or knowing the specific model formulas. NDCN is proposed to use the Ordinary Differential Equation system combined with the Graph Neural Network block to learn the continuous network dynamics process~\cite{zang2020neural}. The GNN block is used to model the complicated and nonlinear interactions between node states, and the real-time dynamic changes of node states are caught or tracked by the ODE system. The real-time states of nodes can be achieved similarly by solving an ODE initial-value problem. The depth of GNN blocks is analogous to the length in the time domain. NDCN uses a single ODE system to model the network dynamics, and the predicted states of nodes are obtained by solving the same ODE initial-value problem. However, using a single ODESolver, NDCN can not handle the long-term network dynamics well. The deviation between the predicted values and real states of nodes may get larger and larger over time if the predicted states of nodes in the early period do not fit well with the true observations. In addition, NDCN does not make full use of all the observations effectively. The observations of node states other than the initial values are not engaged in the training process as prior knowledge. This is probably one of the main reasons for this bad performance in the predictions of the long-term network dynamics. And this situation will become even worse when the observations are very sparse. It is difficult for a single ODE system to learn or fit the real dynamics of node states with sparse observations. To mitigate or overcome the effect of the sparsity of observations on NDCN and improve the ability of the long-term prediction, instead of directly using ODE systems to model the entire network dynamics process like NDCN, we model the network dynamics in segments using the ODE equations in an autoregressive way, which can reduce the difficulty of learning the dynamic process. In this way, the true observations can be fully involved in the training process as prior knowledge and effectively correct the prediction bias at each timestamp to ensure that the initial value for the next prediction is as accurate as possible. By this means, we impose a strong constraint on both ends of the ODESolver to reduce the prediction error and force the ODE system to easily capture or fit the real dynamics. 

In this paper, we propose an \textbf{A}utoregressive \textbf{G}NN-\textbf{O}DE \textbf{G}RU Model(\textbf{AGOG}) to learn and capture the continuous network dynamics in a data-driven manner. Considering the sparsity of the observations of node states, we use the Autoregressive model to make a one-step ahead prediction based on the observation history to guarantee the continuity and accuracy of node states. The GNN module is utilized to model the complicated and nonlinear network dynamics, and this module includes two parts that imply the interactions between nodes and the dependence of dynamic evolution, respectively. The dynamics of hidden states of node states is formulated by the ODE system. The augmented ODE system~\cite{dupont2019augmented} is used to map the hidden state of node states into the continuous time domain, and the hidden state is solved in the expanded space, which could improve generalization and stability and achieve lower losses. And the hidden states are updated through observations by GRUCell. The objective function is defined to minimize the reconstruction error and continuity error which describe the accuracy of prediction and the continuity of dynamic processes, respectively. To evaluate the performance of our model, in consideration of these three tasks: interpolation reconstruction, extrapolation prediction, and regular sequences prediction, we visualize the learned network dynamics and test our model and other baseline methods in different network dynamics with various underlying networks. The results demonstrate that our model can learn and capture the continuous network dynamics precisely and stably. Our model consistently outperforms other baseline methods in most cases and can still achieve comparable performance in other cases.

The rest of this paper is organized as follows. The following Section is the Related Work. Our model is described in Section $3$. The experiments are executed in Sections $4$ and $5$. This paper is summarized in the last Section.

\section{Related work}
The proposal of Neural ODE~\cite{chen2018neural} causes a sensation and ignites the enthusiasm of researchers. Different from discrete hidden layers in the neural networks, Neural ODE uses the ODE system as a fundamental component, and the output is calculated through the ODESolver. The continuous neural dynamics of Neural ODE determines that it can handle or incorporate time-series data, especially with non-uniform intervals, which inspires further research greatly~\cite{morrill2020neural,jhin2021ace,ijcai2021-207}. Rubanova et al.~\cite{rubanova2019latent} define two models: autoregressive ODE-RNN and Latent ODE based on variational auto-encode, to handle the time series with irregular intervals. Latent ODE uses ODE-RNN as the recognition network to identify the approximate posterior of the initial state, and the ODESolver is then utilized to achieve the hidden state at any time point. Lechner et al.~\cite{lechner2020learning} theoretically prove that ODE-RNN is plagued by gradient vanishing and gradient explosion and introduce ODE-LSTM to address this problem and capture the long-term dependency in irregularly sampled time-series data. Inspired by the Neural ODE, Brouwer et al.~\cite{brouwer2019gru} modify the Gated Recurrent Unit in the continuous time domain and combine it with the Bayesian update network to model the sporadic observations. Herrera et al.~\cite{herrera2020neural} consider the lack of theoretical guarantee for the predictive capabilities of methods mentioned above and then propose Neural Jump ODE to model the conditional expectation for stochastic processes in continuous time.

The powerful capabilities of Graph Neural Networks(GNN) have been proved in various tasks, such as node classification, link prediction, and graph classification~\cite{zhang2020deep,wu2020comprehensive}. In recent years, multiple variants of GNN spring up such as GCN~\cite{kipf2016semi}, GAE~\cite{kipf2016variational}, GAT~\cite{veli2018graph} and GraphSage~\cite{hamilton2017inductive}. GCN~\cite{kipf2016semi} can be divided into two classes: spectral-based and spatial-based approaches. Spectral-based approaches define the convolutional operator in the spectral domain, and the convolutional operator is deemed as the noise filter in the graph signal processing~\cite{li2018adaptive}. Researchers consider designing different convolutional operators to capture different structural properties of graphs. Spatial-based approaches define message passing rules or convolutional operators based on graph topology, and the node representation or node message is updated by its neighbors' and its own~\cite{zhuang2018dual,gilmer2017neural}. To improve the performance of GNN, some tricks or modules are proposed. Skip connection is tried to alleviate the over-smoothing and make GNN deeper~\cite{li2019deepgcns}. Sampling methods consider reducing the size of node neighbors by conducting sampling for each node or each convolution layer~\cite{chen2018fastgcn}. Pooling methods focus on designing pooling layers to learn high order or graph level hierarchy representation of graphs~\cite{ying2018hierarchical}. Besides, GNN combined with gate mechanisms is also proposed to use LSTM or GRU in the message passing to improve the long-term passing of node information~\cite{li2016gated}.

The work of combining GNN and Neural ODE or PDE has also been proposed. NDCN~\cite{zang2020neural} uses the GNN-based ODE system to learn the nonlinear and complex continuous dynamics on the network and achieve predictions of node states at an arbitrary time. The GNN module is utilized to model the network dynamics equation, and the ODE system is used to map the discrete node states into the continuous time domain. Besides, NDCN is also suitable for the node classification task. The continuous dynamics of ODE corresponds to the discrete layers of GNN, and the continuous time can be interpreted as the depth of discrete layers. Similarly, CGNNs~\cite{xhonneux2020continuous} generalize GNN with discrete layers into a continuous physical process. Inspired by the idea of Pagerank and graph diffusion methods, CGNNs use the ODE system to define the continuous dynamics of node embedding, and the derivative of node embedding is derived from the embedding of its neighbors and itself and its initial values. CGNNs can effectively improve over-smoothing, help make GNN deeper, and learn long-term dependency. Grand~\cite{chamberlain2021grand} is also motivated by the continuous diffusion process and model graph learning as a discretization of a basic PDE. And Grand uses different spatial and temporal discretization formalism and diffusivity functions to model the diffusion on the graph. However, CGNNs and Grand are both derived from the graph diffusion mechanisms, designed for the node classification task, and not particularly targeted for modeling the dynamic processes between nodes, and the terminal states or the final representation of nodes are achieved in a short time. Moreover, STGODE~\cite{fang2021spatial} applies CGNN to the Traﬀic Flow Forecasting. STGODE uses the continuous GNN-ODE to capture the spatial-temporal dependency and alleviate over-smoothing.

\section{Autoregressive GNN-ODE GRU Model}

Consider that a continuous dynamic process on a network is governed by:
\begin{equation}
 \frac{dx(t)}{dt} = f(x,G,t),
\end{equation}  
where $x(t)$ are the states of nodes and $G=(V,E)$ represents the underlying network structure that describes the interaction relationships between nodes. And $f$ is the governing function which dominates the dynamic law of the evolution of node states. The number of nodes of $G$ is $n$, and $x(t) \in \mathbb{R}^{n \times k}$ where $k$ denotes the dimension of node states. And $k$ can be any integer greater than zero. 

\begin{figure*}[!t] 
\centering 
\includegraphics[width=0.9\textwidth]{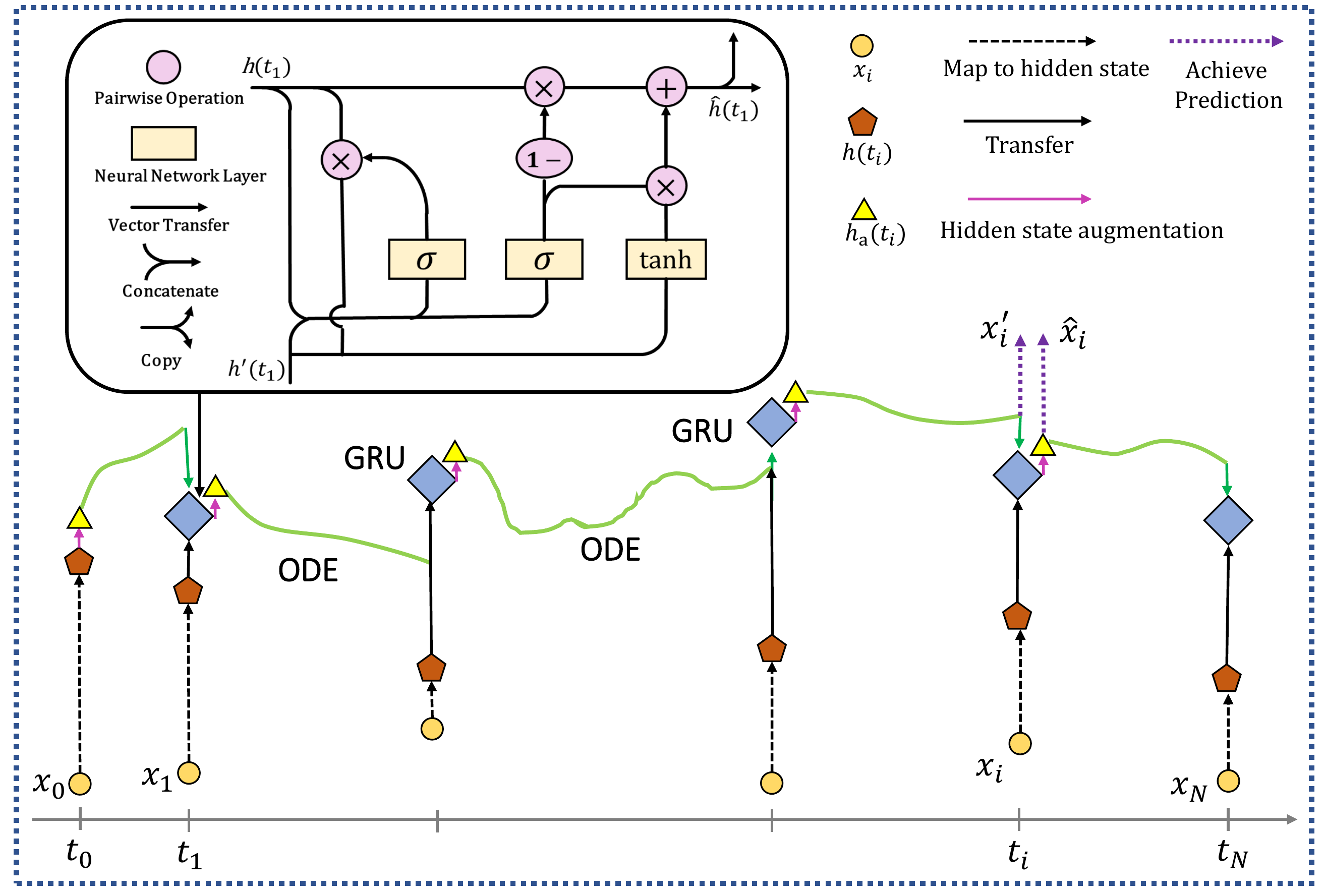} 
\caption{Diagram of Autoregressive GNN-ODE GRU Model. The process of GRUCell\cite{daoulas2021open} is presented inside the black box. Our model provides one-step ahead predictions based on observations history in the form of an autoregressive model. The observation is first mapped into the hidden state. And the dynamic of hidden states is formulated by the Augmented ODE system. The hidden state at the next moment is achieved by solving an ODE initial value problem through the ODESolver. Predictions of node states are obtained from the hidden state. Then the hidden states are updated through GRUCell with new observations at the same time as the prior knowledge. The updated hidden states are used to predict the node states at the next moment. Procedures are repeated over and over to achieve predictions of node states at the corresponding time.}
\label{diagram} 
\end{figure*}

For this continuous dynamic process on the network, we can have a series of observations of node states, $\mathbf{x} = \{(t_0, x_0),(t_1, x_1),...,(t_i, x_i),...,(t_T, x_T)\}$, where each $t_i \in \mathbb{R}$ denotes the timestamp of the observation $x_i \in \mathbb{R}^{n \times k}$ and $t_0<...<t_i<...<t_T$. We want to realize predictions of the node states on $G$ at any given time which may be larger or smaller than $t_T$. To tackle this problem, we propose an Autoregressive GNN-ODE GRU model to provide a novel solution. Autoregressive models make further predictions based on the observation history and specify the conditional probability $p(x_i|x_i-1,...,x_0)$. One-step ahead predictions are achieved by solving an ODE initial-value problem. And the new observations as prior knowledge are combined with the hidden states for the predictions at the next timestamp. The diagram of our model is shown in Fig. \ref{diagram}, and the framework of our model is specified as follows.

Suppose that the initial observation of node states is $x_0 \in \mathbb{R}^{n \times k}$, and the corresponding observation timestamp is $t_0$. We first map $x_0$ into a hidden state $h(t_0)\in \mathbb{R}^{n \times d}$:

\begin{equation}
 h(t_0) = x_0 W_e + b_e,
\end{equation} 
in which $W_e \in \mathbb{R}^{k \times d}$ and $b_e \in \mathbb{R}^{n \times d}$.

The dynamic of hidden states is specified by the following ordinary differential equation system:

\begin{equation}
\label{gcn_ode}
 \frac{dh(t)}{dt} = g_\theta(h(t),t),
\end{equation}     
in which the function $g_\theta(h(t),t)$ formulates the latent dynamic of hidden states with the parameter $\theta$. Consider that the dynamic process is modeled or relies on the network structure. The function $g_\theta$ is designated expressly by a GNN module:
\begin{equation}
\label{gcn}
 g_\theta = \Phi h(t) \theta_1 + (h(t) \theta_0 +b_0),
\end{equation}
where $\Phi$ is the normalized Laplacian matrix, $\Phi = D^{-\frac{1}{2}} (D-A) D^{-\frac{1}{2}}$ in which $A$ denotes the adjacency matrix of $G$ and $D$ denotes the degree matrix of the network. On the right side of Eq. \ref{gcn}, the left part models the interactions of hidden states between nodes, while the right part specifies the dependence of the dynamic evolution of the hidden state on the current hidden state of nodes. The left part simulates the influence of dynamic interactions between nodes on the state of nodes, and the right part represents the self-dynamic of nodes which implies the evolution rule of the node states influenced by their own states.

So, the next hidden state at $t_1$, $h^{\prime}(t_1)$, can be achieved by the ODESolver:

\begin{gather}
h^{\prime}(t_1) = h(t_0) + \int_{t_0}^{t_1} g_\theta(h(t),t)\, dt, \\
h^{\prime}(t_1) = ODESolver(g_\theta(h(t),t), h(t_1), (t_0, t_1)).  
\end{gather} 
This initial-value problem is solved by numerical methods, e.g., the fourth-order Runge-Kutta method, the high-order Dormand-Prince DOPRI5 method, and the Euler method. We use the ODE solvers from the torchdiffeq package (https://github.com/rtqichen/torchdiffeq), and details can be found in ~\cite{chen2018neural}.
To make the ODESolver more powerful and expressive, we adopt the Augmented Neural ODE~\cite{dupont2019augmented} to offer a novel solution to the problems above. The space of the hidden state is expanded, and the ODE is solved from $\mathbb{R}^{n \times d}$ to $\mathbb{R}^{n \times (d+p)}$ where $p$ represents the augmented dimension. The augmented space is defined as $a(t) \in \mathbb{R}^{n \times p}$, and Eq. \ref{gcn_ode} is updated to:

\begin{equation}
 \frac{d[h(t), a(t)]}{dt} = g_\theta([h(t), a(t)], t), [h(0), a(0)] = [h(t_0), \mathbf{0}].
\end{equation}
For simplicity, the hidden state with the augmented space is denoted as $h_a(t)$, and $h_a(t_0)=[h(t_0), \mathbf{0}]$. Intuitively, the hidden state $h(t)$ is merged with the augmented zero space $a(t)$, and the ODE is solved in the expanded space. With the additional space, the ODE flow can easily flow to the augmented space, and the learned hidden state is represented in a higher dimension and has a better expressivity and precision. Simultaneously, the trained function $G_\theta$ becomes more smooth and achieves better generalization and lower time cost. Then the predicted hidden state $h^{\prime}(t_1) \in \mathbb{R}^{n \times (d+p)}$ at $t_1$ is achieved by:
\begin{equation}
h^{\prime}(t_1) = ODESolver(g_\theta, h_a(t_0), (t_0, t_1)).  
\end{equation}
And the prediction of node states at $t_1$, $x^{\prime}_1$, is achieved through a linear layer:
\begin{equation}
\label{output1}
x^{\prime}_1 = h^{\prime}(t_1) W_o+ b_o, 
\end{equation}
where $W_o \in \mathbb{R}^{(d+p) \times k}$ and $b_o \in \mathbb{R}^{n \times k}$.

The real node states at $t_1$ are denoted as $x_1$. We use real observations as the prior knowledge to revise and update the obtained hidden state, $h^{\prime}(t_1) \in \mathbb{R}^{n \times (d+p)}$. We also map $x_1$ into the hidden state as $h(t_1) \in \mathbb{R}^{n \times d}$ and utilize the GRUCell to update hidden state $h^{\prime}(t_1)$ at $t_1$ given current true observation $x_1$:
\begin{gather}
h(t_1) = x_1 W_e + b_e,\\
\hat{h}(t_1) = GRUCell(h^{\prime}(t_1),h(t_1)). \label{gru}
\end{gather}
In detail, Eq. \ref{gru} is formulated as:
\begin{equation}
\begin{split}
r_{1} =\sigma( h^{\prime}(t_1)W_{r} + b_{wr} + h(t_1)U_{r} +b_{ur}),\\
z_{1} =\sigma( h^{\prime}(t_1)W_{z} + b_{wz} + h(t_1)U_{z} +b_{uz}),\\
n_{1} =\tanh ( h^{\prime}(t_1)W_{h} + b_{wh} + r *( h(t_1)U_{h}+b_{uh})),\\
\hat{h}(t_1) =(1-z_{1}) * n_{1}+z_{1} * h(t_1),
\end{split}
\end{equation}
in which $W_{h} \in \mathbb{R}^{(d+p) \times d}$, $U_{h} \in \mathbb{R}^{d \times d}$, $b_{wh}, b_{uh} \in \mathbb{R}^{n \times d}$. The updated hidden state at $t_1$ is denoted as $\hat{h}(t_1) \in \mathbb{R}^{n \times d}$. The augmented hidden state $h_a(t_1)$ at $t_1$ is expressed as $[\hat{h}(t_1), \mathbf{0}]$, and the updated prediction of node states at $t_1$ from the augmented updated hidden state $h_a(t_1)$, can be calculated from:
\begin{equation}
\label{output2}
\hat{x}_1 = h_a(t_1) W_o+ b_o.  
\end{equation}
Then $h_a(t_1)$ can be the initial value of the ODESolver to get the hidden state at the next moment $t_2$, $h^{\prime}(t_2)$:
\begin{equation}
h^{\prime}(t_2) = ODESolve(g_\theta, h_a(t_1), (t_1, t_2)).  
\end{equation}

And the prediction $x^{\prime}_2$ at $t_2$ can be achieved through $h^{\prime}(t_2)$. Repeating these procedures above, we can obtain the prediction of node states at the corresponding time. There are two types of outputs, $x^{\prime}_i$ and $\hat{x}_i$, defined as the prediction and the updated prediction to show the distinction, respectively. $x^{\prime}_i$ is obtained from the solution of the ODESolver $h^{\prime}(t_i)$, through a linear layer based on Eq. \ref{output1}, while $\hat{x}_i$ is achieved from the augmented updated hidden state $h_a(t_i)$, through a linear layer based on Eq. \ref{output2}. The augmented updated hidden state $h_a(t_i)$ is obtained in two steps. The predicted hidden state of the ODESolver solution $h^{\prime}(t_i)$ is updated by the GRUCell through the real observations $x_i$ at the same time to get the updated hidden states $\hat{h}(t_i)$, then $h_a(t_i)$ is achieved after the augmentation of the $\hat{h}(t_i)$. And $x^{\prime}_i$ is deemed as the final result that we need.

The loss function is defined as below:

\begin{equation}
\label{loss}
L = \frac {1}{T}\sum_{i= 0}^T|x^{\prime}_i - x_i| + \frac {1}{T}\sum_{i=0}^T|\hat{x}_i - x^{\prime}_i|,
\end{equation}
where $| \cdot |$ denotes the $l_1$-normal loss, that is the mean element-wise absolute value difference. Our aim is to minimize this objective function, Eq. \ref{loss}. The above function has two terms: the left part is the reconstruction error, while the right part is the continuity error. It is easy to understand the reconstruction error which requires the predicted value to be closer to the true states. As for the continuity error, in the dynamic of the hidden states, we expect that the updated hidden states should be consistent with the predicted hidden states from the ODESolver at the same timestamp. The updated hidden states are then used as the initial value for the predictions at the next timestamp. We use the true observations to correct the hidden state and hope that the updated hidden state will not deviate too much from the original hidden state. By introducing the true observations, we impose a strong constraint on both ends of the ODESolver and force the ODE system to capture the real dynamics easily. Moreover, the continuity error implies the constraints on the smoothness and continuity of the trajectory of the predicted states. $W_e, W_o, \theta_0, \theta_1, b_e, b_o, b_0$, and GRUCell are all trainable parameters, and they are updated through backpropagation of the gradient of the loss. The loss function is optimized by the Adam method, and the pseudocode of our Autoregressive GNN-ODE GRU model is presented in Algorithm \ref{algorithm}. For clarity, our model is abbreviated as \textbf{AGOG}(\textbf{A}utoregressive \textbf{G}NN-\textbf{O}DE \textbf{G}RU Model).

\begin{algorithm}[H]
  \caption{Autoregressive GNN-ODE GRU Model} 
  \label{algorithm}
   \begin{algorithmic}[1]
   \renewcommand{\algorithmicrequire}{\textbf{Input:}}
\Require
augmented dimension: $p$; number of nodes: $n$; node states and their timestamps: $\mathbf{x} = \{(t_i, x_i)\}_{i=0...T}$;
\renewcommand{\algorithmicrequire}{ \textbf{Output:}}
    \Require
      predicted node states: $\{x^{\prime}_i\}_{i=0...T}$ and $\{\hat{x}_i\}_{i=0...T}$;

\For{$i = 0 : T$}
        \State $h(t_i) = x_i W_e + b_e$
    \EndFor
    \State $h_a(t_0)=[h(t_0), \mathbf{0} \in \mathbb{R}^{n \times p}]$
    \State $h^{\prime}(t_1) = ODESolve(g_\theta, h_a(t_0), (t_0, t_1))$
    \State $\hat{h}(t_1) = GRUCell(h^{\prime}(t_1),h(t_1))$
    \For{$i = 2 : T$}
            \State $h_a(t_{i-1})=[\hat{h}(t_{i-1}), \mathbf{0} \in \mathbb{R}^{n \times p}]$
            \State $h^{\prime}(t_i) = ODESolve(g_\theta, h_a(t_{i-1}), (t_{i-1}, t_i))$
            \State $\hat{h}(t_i) = GRUCell(h^{\prime}(t_i),h(t_i))$
    \EndFor
    \For{$i = 0 : T$}
        \State $x^{\prime}_i = h^{\prime}(t_i) W_o+ b_o$
        \State $\hat{x}_i = h_a(t_i) W_o+ b_o$
    \EndFor
   \end{algorithmic}
\end{algorithm}

\section{Interpolation reconstruction and Extrapolation prediction}

We evaluate our model on different structures of networks with various network dynamics. Complex network models include: 1) Community network~\cite{fortunato2010community}, 2) Grid network, 3) Erdós and Rényi random network~\cite{erd6s1959random}, 4) Power-law network~\cite{barabasi1999emergence}, and 5) WS small-world network~\cite{watts1998collective}. For each type of network models, we investigate three kinds of network dynamics processes. And the real benchmark dynamic of node states is generated according to its underlying network structure and rules of dynamic evolution by the Dormand-Prince method. Regarding the experimental setup, we generally follow the settings in NDCN~\cite{zang2020neural}, except for setting a larger time interval for observations, because we consider the sparsity of observations in real life. In the simulation, for each network dynamics on each network structure, we randomly sample 120 snapshots of node states with irregular time interval, $\mathbf{x} = \{(t_0, x_0),(t_1, x_1),..., (t_{119}, x_{119}), t_0<t_1...<t_{119}\}$, where sampling intervals between adjacent datapoints are not equal. The last $20$ datapoints $\{(t_i,x_i)\}_{i=100,...,119}$ are used for the extrapolation prediction task. As for the first 100 datapoints, we randomly choose $P\%$ datapoints for training, and the residual datapoints are targeted for the interpolation reconstruction task. A smaller $P$ means that the observed data is more sparsity. To illustrate the superiority of our model, we use the \textbf{MAE}(Mean Absolute Error), and \textbf{Normalized L1 loss} metrics to measure the performance of our model on these two tasks. The observed node states may have zero, so the MAPE(Mean Absolute Percentage Error) is not applicable here. Instead, the \textbf{Normalized L1 loss} is selected, which is normalized by the mean value of all observations.

\textbf{Experimental configuration}. The number of nodes of the generated networks is set as $400$. The initial node state $x_0$ is set as the same for each case. The training percentage of observed snapshots $P\%$ is set as $10\%$, and $30\%$, respectively. With regard to hyperparameters in our model, the hidden dimension $d$ of hidden state $h(t_i)$ is set as $20$, $h(t_i) \in \mathbb{R}^{n \times 20}$, and the augmented dimension $p$ is set as $5$. And our model is trained for $800$ epochs with the learning rate $0.01$ in all cases. The Euler method is used for the ODESolver. 

We use NDCN method~\cite{zang2020neural} as the baseline to compare with our model. NDCN uses a simple GNN module to model the interactions between node states, and the discrete node states are mapped into the continuous time domain by the ODE system. The main goal of NDCN is state prediction on the network. There are also three variants of NDCN: No-Encode, No-Graph, and No-Control. For succinctness, we only consider NDCN in this paper because of its best performance between it and its variants. Moreover, there are also some GNN-ODE methods that are not designed for predictions of node states, but we consider them as baselines here as well. These methods define the evolution function of node embedding as the ODE function, and the final embedding is achieved by the ODESolver in a finite time. Here we regard the node embedding as the hidden state of nodes states. We use the ODE functions of these GNN-ODE methods to model the network dynamics, respectively. CGNNs~\cite{xhonneux2020continuous} expand the discrete GNN layers into a continuous physical process inspired by the diffusion process on graphs and define an evolution equation of node embedding that the derivative of node embedding is the combination of itself, the embedding of its neighbors, and its initial value. CGNNs have two different variants, which consider whether each dimension of node embedding evolves independently or not, and we denote them as CGNN and WCGNN, respectively, which are all considered here. Similarly, Grand~\cite{chamberlain2021grand} is also motivated by the continuous diffusion process and model GNN as a discretization of a basic PDE. And Grand uses different spatial or temporal discretization formalisms and diffusivity functions to model the diffusion on the graph. The diffusivity is modeled by the scaled dot product attention mechanism. The more general non-linear form of GRAND~\cite{chamberlain2021grand} is tested here, denoted as GRAND. We use the default parameters for NDCN as stated in ~\cite{zang2020neural}. The hidden size is set as $20$ for CGNN, WCGNN, and GRAND. And the dimension of the query and key matrices is both set as $20$ for GRAND. The loss function of CGNN, WCGNN, and GRAND is the same as that of NDCN.

\subsection{Network dynamics model}
We consider several following network dynamics from different domains in this paper. Define the adjacent matrix of the underlying network as $A$ and the state of node $i$ at timestamp $t$ as $z_i(t) \in \mathbb{R}^{1 \times k}$ in which $k$ denotes the dimension of the node state. In the experiment, $k$ is set as $1$ for all network dynamics. The descriptions of these network dynamic models and specific governing equations of dynamics are listed in turn.

\begin{itemize}
\item {\textbf{Gene regulatory network dynamics model}}~\cite{gao2016universal}. This model simulates the interactions between genes which are governed by:

\begin{equation}
 \frac{dz_i(t)}{dt} = -b_i z_i +  \sum_{j=1}^N A_{ij}\frac{z_j^h}{z_j^h+1}.
\end{equation}  

\item {\textbf{Kuramoto network dynamics model}}~\cite{kuramoto1975self}. This model describes the synchronizations of phase-coupled oscillators and is defined as:

\begin{equation}
 \frac{dz_i(t)}{dt} = \omega_i +  k_i \sum_{j=1}^N A_{ij} \sin(z_i - z_j).
\end{equation} 

\item {\textbf{Mutualistic interaction network dynamics model}}~\cite{gao2016universal}. This model describes the mutualistic interactions between species in the ecotope, and the specific dynamics equation is shown as:

\begin{equation}
\begin{split}
 \frac{dz_i(t)}{dt} = -b_i + z_i(1-\frac{z_i}{k_i})(\frac{z_i}{c_i}-1) + \\
 \sum_{j=1}^N A_{ij}\frac{z_i z_j}{d_i+e_i z_i+h_j z_j}.
\end{split}
\end{equation} 

\end{itemize}

\begin{figure*}[!t] 
\centering 
\includegraphics[width=0.995\textwidth]{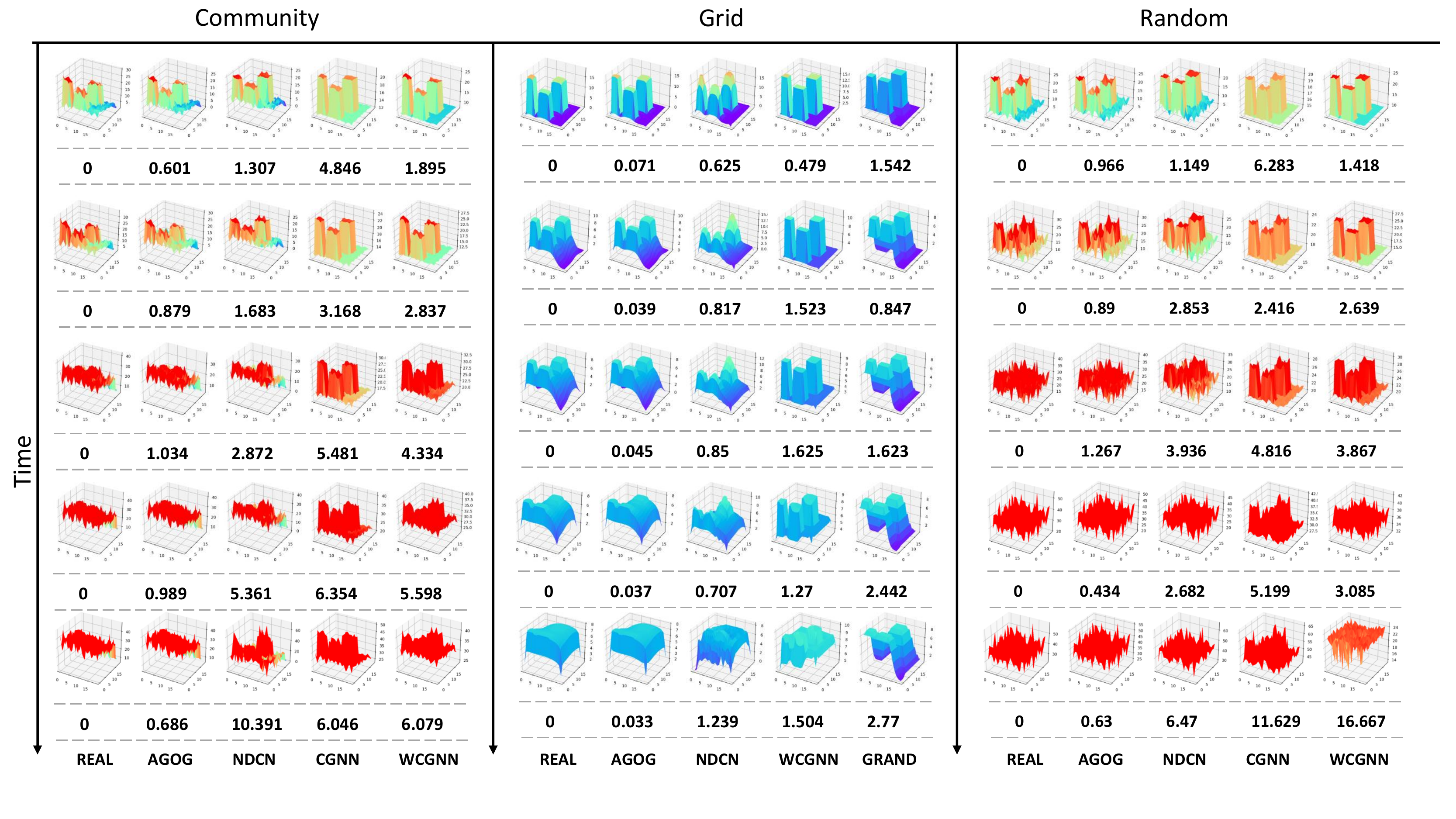} 
\caption{Snapshot visualizations of the Gene regulatory network dynamics in different underlying networks with $P\%$ as $10\%$.} 
\label{Gene}
\end{figure*}

\begin{figure*}[!] 
\centering 
\includegraphics[width=0.995\textwidth]{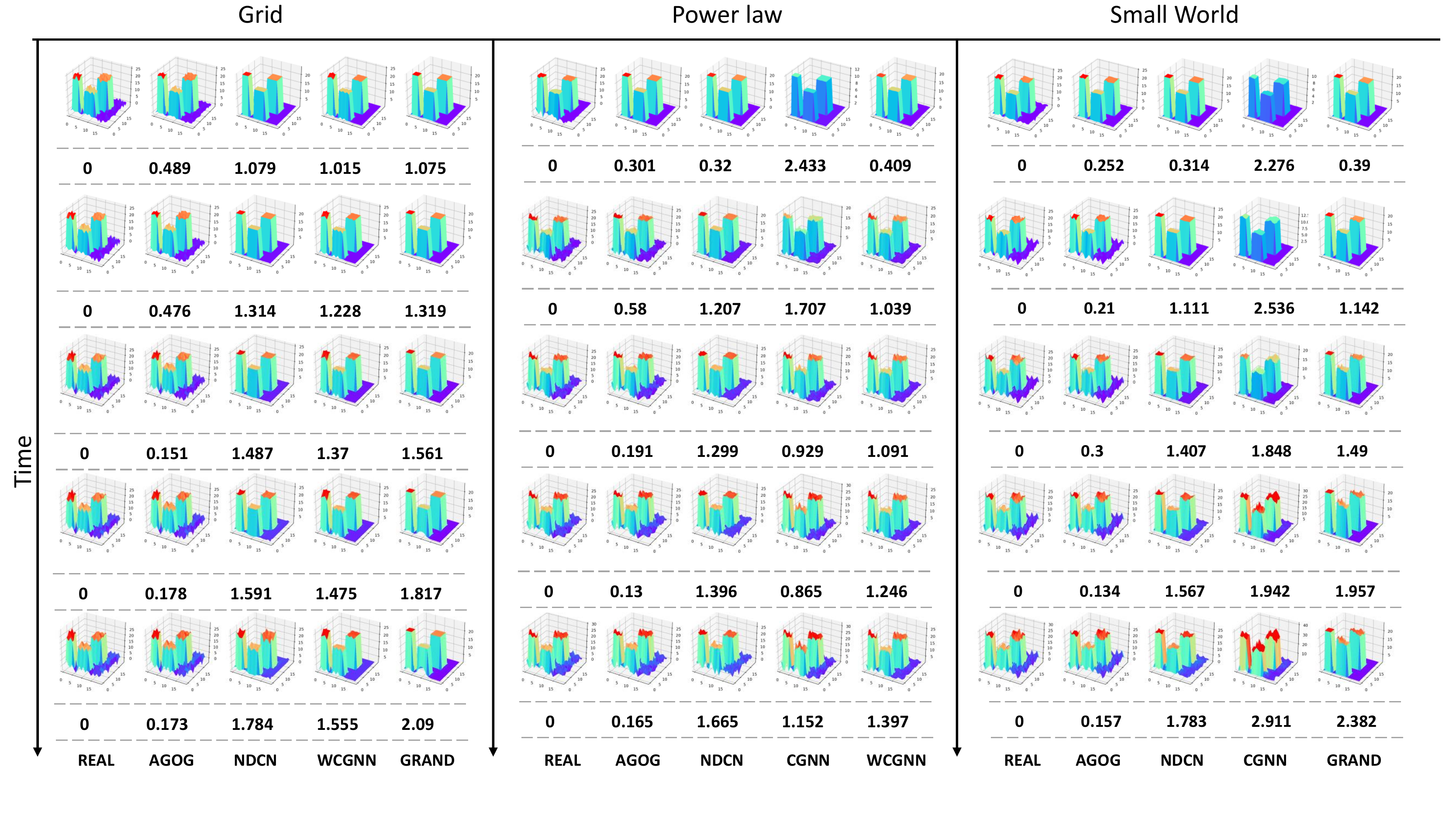} 
\caption{Snapshot visualizations of the Kuramoto network dynamics in different underlying networks with $P\%$ as $10\%$.} 
\label{Kuramoto}
\end{figure*}

\begin{figure*}[!t] 
\centering 
\includegraphics[width=0.995\textwidth]{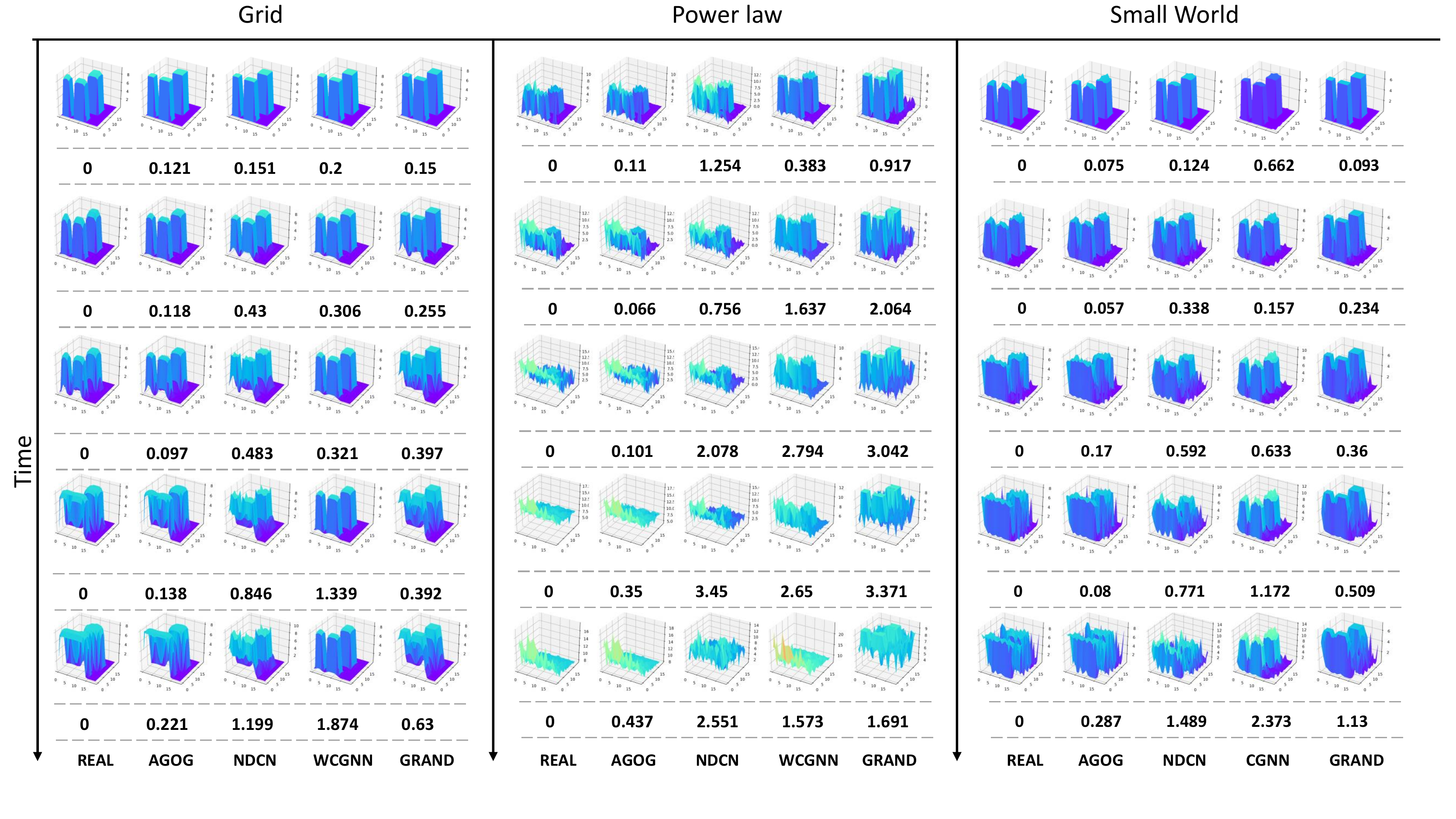} 
\caption{Snapshot visualizations of the Mutualistic interaction network dynamics in different underlying networks with $P\%$ as $10\%$.} 
\label{Mutual}
\end{figure*}

\subsection{Result}

We visualize the snapshots of node states of each network dynamics with various underlying networks learned by different methods in Figs. \ref{Gene}, \ref{Kuramoto}, and \ref{Mutual}, where the training percentage of observed snapshots $P\%$ is set as $10\%$. Due to space limitations, we just present the dynamic process on three networks for each network dynamics. In each case, we select five snapshots of node states, and node states evolve along the direction of time from top to bottom. In each snapshot, $400$ nodes are mapped into a $20 \times 20$ matrix, and each square corresponds to a node and reflects its state value. In each row of each case, the five photos show the real dynamic of node states, the dynamic captured by our model AGOG, and the dynamic learned by NDCN and the other two selected methods in turn, from left to right. The number at the bottom of each row of images represents the difference between the true value and prediction of the node state by different methods, which is quantified by the MAE metric. (In addition, we also show the quantitative results over time of the interpolation reconstruction and extrapolation prediction tasks in the Figs. \ref{Gene_over_time}, \ref{Kuramoto_over_time} and \ref{Mutual_over_time})

As shown in these figures, each type of network dynamics with different structures of networks performs various evolution processes of node states except the Kuramoto network dynamics. This may be because the Kuramoto network dynamic is a long-term dynamic process. The result demonstrates that our model can learn and capture the network dynamics accurately in most instances. In rare cases, our model exhibits a slight time-lag but can still catch the real dynamics in a very short time. Moreover, the quantitative results also show that our method can maintain an accurate prediction of node states with a small error at any arbitrary time. But it is hard for these baselines to capture the real dynamics. It is clear that these methods either do not learn the network dynamics well from the beginning or their predictions deviate more and more from the true states as time goes on. For example, in some cases, NDCN can learn the pattern of the real dynamics initially, but the value of predicted node states quickly contrasts sharply with the real one, which is more obvious when the dynamic evolves for quite a long time. It may be a challenge of robustness or the learning ability of the long-term for NDCN. And NDCN exhibits severe time delay in most cases. At the same time, the sparsity of the observations also causes great difficulties for NDCN. As for other GNN-ODE based methods, it is very tough for them to capture the real evolution of network dynamics, especially for GRAND. WCGNN has a bright performance in some scenes, but at the same time, it shows a strong time lag.

The experiments are conducted over 20 independent realizations, and the average performance in all cases is presented in Tabs. \ref{inter_0.1} and \ref{extra_0.1}, corresponding to the interpolation reconstruction task and extrapolation prediction task, respectively(owing to space constraints, results with the percentage of observed node states as $30\%$ are shown in the APPENDICES). In each row, the results of our model AGOG and other baseline methods under these two metrics in a certain case are shown from left to right, respectively. We also present the results of our method without considering the continuity loss in the objective function, which is abbreviated as AGOG*. Each result is reported after multiplying by $10^{-2}$, and the best result in each case and each metric is in bold. The results show that our model outperforms other baseline methods in both interpolation reconstruction and extrapolation prediction tasks by a huge margin, especially when the proportion of the observed node states is very small. NDCN cannot handle very sparse data well and learn the network dynamics from that. Our model presents considerable robustness and can learn and simulate the continuous network dynamics precisely. The accuracy of our model is one order of magnitude higher than other baseline methods in most cases. In the gene regulation dynamics, the Normalized L1 metric of our model can be $9.5\%$ of that of NDCN and $10.3\%$ of that of WCGNN in the extrapolation prediction task, and $20.4\%$ of that of NDCN and $18.9\%$ of that of WCGNN in the interpolation reconstruction on average. And in the Mutualistic interaction network dynamics, the average MAE metric accuracy of our model can achieve $14.1 \times 10^{-2}$ in the extrapolation prediction task, which is $9.7\%$ of that of WCGNN. These two metrics reach the same conclusion, and all demonstrate the superiority and unparalleled ability of our model. Moreover, with the introduction of the continuity loss, the performance of AGOG is significantly improved in comparison with AGOG*, in both extrapolation and interpolation prediction tasks. The continuity of the trajectory of the predicted nodes states contributes to the accuracy of predictions. In rare cases, AGOG performs slightly worse than AGOG*. Under these conditions, the introduction of continuity loss may cause over-smoothing, making the true observations less useful. The balance between the reconstruction error and the continuity error may be a good choice to alleviate the over-smoothing.

\begin{table*}[]
\footnotesize
\caption{Results($\times 10^{-2}$) of interpolation reconstruction of network dynamics with the percentage of observed node states as $10\%$}
\label{inter_0.1}
\centering
\begin{tabular}{cccccccccccccc}
\hline
                                                                                         &             & \multicolumn{6}{c}{MAE}                                               & \multicolumn{6}{c}{Normalized   L1}                            \\ \hline
                                                                                         &             & AGOG            & AGOG*           & NDCN   & CGNN   & WCGNN  & GRAND  & AGOG           & AGOG*         & NDCN  & CGNN  & WCGNN & GRAND \\ \hline
\multirow{5}{*}{\begin{tabular}[c]{@{}c@{}}Gene   \\      Regulation\end{tabular}}       & Community   & \textbf{83.36}  & 85.21           & 422.7  & 569.22 & 459.95 & 624.1  & \textbf{3.38}  & 3.45          & 17.14 & 23.06 & 18.63 & 25.35 \\
                                                                                         & Grid        & \textbf{12.05}  & 20.4            & 104.81 & 205.01 & 121.81 & 165.95 & \textbf{2.65}  & 4.5           & 23.06 & 45.18 & 26.81 & 36.49 \\
                                                                                         & Power Law   & \textbf{33.33}  & 37.24           & 114.34 & 310.83 & 226.75 & 332.87 & \textbf{4.27}  & 4.78          & 14.66 & 39.87 & 29.06 & 42.67 \\
                                                                                         & Random      & \textbf{100.97} & 134.49          & 541.98 & 582.36 & 357.37 & 744.93 & \textbf{3.32}  & 4.44          & 17.7  & 18.99 & 11.63 & 24.34 \\
                                                                                         & Small World & \textbf{17.19}  & 29.16           & 75.05  & 156.04 & 73.77  & 127.79 & \textbf{4.07}  & 6.92          & 17.69 & 36.81 & 17.38 & 30.16 \\ \hline
\multirow{5}{*}{\begin{tabular}[c]{@{}c@{}}Kuramoto \\      Model\end{tabular}}          & Community   & \textbf{30.33}  & 32.24           & 148.65 & 152.8  & 106.95 & 175.32 & \textbf{6.28}  & 6.67          & 30.77 & 31.64 & 22.14 & 36.29 \\
                                                                                         & Grid        & \textbf{27.86}  & \textbf{27.07}  & 145.01 & 223.28 & 135.24 & 157.22 & \textbf{5.74}  & \textbf{5.57} & 29.86 & 46    & 27.86 & 32.38 \\
                                                                                         & Power Law   & \textbf{25.02}  & 26.11           & 119.98 & 133.18 & 88.06  & 176.38 & \textbf{5.12}  & 5.35          & 24.56 & 27.28 & 18.03 & 36.11 \\
                                                                                         & Random      & \textbf{32.51}  & \textbf{32.02}  & 154.12 & 139.67 & 103.79 & 174.01 & \textbf{6.75}  & \textbf{6.64} & 32.01 & 29.02 & 21.56 & 36.15 \\
                                                                                         & Small World & \textbf{25.84}  & 26.32           & 140.57 & 195.82 & 120.43 & 156.25 & \textbf{5.3}   & 5.39          & 28.8  & 40.14 & 24.67 & 32.01 \\ \hline
\multirow{5}{*}{\begin{tabular}[c]{@{}c@{}}Mutualistic \\      Interaction\end{tabular}} & Community   & \textbf{92.9}   & 105.15          & 263.16 & 345.62 & 235.26 & 344.86 & \textbf{8.34}  & 9.42          & 23.54 & 30.78 & 20.98 & 30.34 \\
                                                                                         & Grid        & \textbf{21.86}  & 23.07           & 51.33  & 88.23  & 56.98  & 36.92  & \textbf{10.27} & 10.83         & 23.94 & 41.18 & 26.55 & 17.24 \\
                                                                                         & Power Law   & \textbf{38.23}  & 62.31           & 181.83 & 188.48 & 164.55 & 208.03 & \textbf{7.23}  & 11.84         & 34.54 & 35.84 & 31.18 & 39.62 \\
                                                                                         & Random      & \textbf{130.9}  & \textbf{110.63} & 303.67 & 321.14 & 237.04 & 301.54 & \textbf{10.15} & \textbf{8.58} & 23.51 & 24.83 & 18.33 & 23.3  \\
                                                                                         & Small World & \textbf{14.85}  & 17.85           & 38.29  & 53     & 29.49  & 25.15  & \textbf{8.79}  & 10.54         & 22.56 & 31.17 & 17.3  & 14.76 \\ \hline
\end{tabular}
\end{table*}

\begin{table*}[]
\footnotesize
\caption{Results($\times 10^{-2}$) of extrapolation reconstruction of network dynamics with the percentage of observed node states as $10\%$}
\label{extra_0.1}
\begin{tabular}{cccccccccccccc}
\hline
                                                                                         &             & \multicolumn{6}{c}{MAE}                                             & \multicolumn{6}{c}{Normalized   L1}                           \\ \hline
                                                                                         &             & AGOG           & AGOG*          & NDCN   & CGNN   & WCGNN  & GRAND  & AGOG          & AGOG*         & NDCN  & CGNN  & WCGNN & GRAND \\ \hline
\multirow{5}{*}{\begin{tabular}[c]{@{}c@{}}Gene \\      Regulation\end{tabular}}         & Community   & \textbf{65.53} & 79.45          & 784.97 & 932.9  & 661.16 & 651.7  & \textbf{2.09} & 2.54          & 25.01 & 29.79 & 21.1  & 20.8  \\
                                                                                         & Grid        & \textbf{8.83}  & 12.39          & 224.64 & 213.69 & 152.79 & 255.68 & \textbf{1.29} & 1.81          & 32.85 & 31.25 & 22.35 & 37.4  \\
                                                                                         & Power Law   & \textbf{26.96} & 36.8           & 152.95 & 426.55 & 288.29 & 428.87 & \textbf{2.8}  & 3.82          & 15.89 & 44.31 & 29.95 & 44.55 \\
                                                                                         & Random      & \textbf{72.87} & 86.45          & 853.83 & 901.31 & 516.93 & 592.88 & \textbf{1.85} & 2.19          & 21.63 & 22.86 & 13.12 & 15.04 \\
                                                                                         & Small World & \textbf{11.1}  & 12.74          & 124.01 & 189.88 & 90.18  & 124.71 & \textbf{1.96} & 2.24          & 21.78 & 33.43 & 15.89 & 21.98 \\ \hline
\multirow{5}{*}{\begin{tabular}[c]{@{}c@{}}Kuramoto \\      Model\end{tabular}}          & Community   & \textbf{13.73} & 17.86          & 210.39 & 166.95 & 114.41 & 218.28 & \textbf{2.26} & 2.94          & 34.72 & 27.5  & 18.86 & 35.97 \\
                                                                                         & Grid        & \textbf{14.94} & 15.58          & 200.52 & 290.94 & 158.56 & 212.69 & \textbf{2.46} & 2.57          & 33.09 & 48.03 & 26.18 & 35.11 \\
                                                                                         & Power Law   & \textbf{13.27} & 15.66          & 190.7  & 156.64 & 108.09 & 244.54 & \textbf{2.18} & 2.57          & 31.31 & 25.68 & 17.71 & 40.09 \\
                                                                                         & Random      & \textbf{13.71} & 15.96          & 185.17 & 135.35 & 104.81 & 214.2  & \textbf{2.26} & 2.64          & 30.7  & 22.5  & 17.38 & 35.51 \\
                                                                                         & Small World & \textbf{13.64} & 14.98          & 194.18 & 245.64 & 141.81 & 214.89 & \textbf{2.23} & 2.45          & 31.81 & 40.27 & 23.24 & 35.21 \\ \hline
\multirow{5}{*}{\begin{tabular}[c]{@{}c@{}}Mutualistic \\      Interaction\end{tabular}} & Community   & \textbf{16.2}  & 23.77          & 194.74 & 316.53 & 179.83 & 205.35 & \textbf{1.17} & 1.71          & 14.01 & 22.78 & 12.95 & 14.81 \\
                                                                                         & Grid        & \textbf{19.94} & 21.04          & 153.61 & 248.75 & 212    & 76.6   & \textbf{5.16} & 5.44          & 39.63 & 64.13 & 54.58 & 19.74 \\
                                                                                         & Power Law   & \textbf{14.9}  & 42.73          & 269.03 & 237.97 & 196.13 & 132.21 & \textbf{1.69} & 4.84          & 30.49 & 26.98 & 22.23 & 14.99 \\
                                                                                         & Random      & \textbf{18.95} & \textbf{17.09} & 240.46 & 253.33 & 168.53 & 142.82 & \textbf{1.23} & \textbf{1.11} & 15.61 & 16.44 & 10.93 & 9.27  \\
                                                                                         & Small World & \textbf{15.9}  & 26.46          & 187.46 & 162.67 & 148.34 & 89.77  & \textbf{5}    & 8.26          & 58.92 & 50.82 & 46.35 & 28.08 \\ \hline
\end{tabular}
\end{table*}

The results of both the snapshots visualizations and quantitative tasks indicate that the sparsity of datapoints poses large difficulties for these baseline methods. From the results of snapshot visualizations, we can clearly find that these algorithms either fail to learn the network dynamics well from the beginning or their predictions deviate more and more from the true state over time. Consider that these baseline methods all use a single ODE function to directly model the overall continuous network dynamics process. And the long-term predictions of methods are based on the answers of the short-term. At the same time, the sparsity of the observations poses great challenges and difficulties for these methods to fit and model these time-series data. Under this condition, once the method fails to predict the short-term data, the prediction bias for the subsequent data will become so large that the network dynamics cannot be well learned. However, we use the autoregressive model to deal with the sparsity of data. We use the ODE function for segmental modeling of the network dynamics to mitigate the problems caused by the sparsity of the data. And the observations are also introduced into the training process at the same time to correct the predicted values at this moment, so that the subsequent predictions are less likely to be affected by the deviations of the previous predictions. In this way, we impose a strong constraint on both ends of the ODESolver and force the ODE system to capture or fit the real dynamics easily. The results show the validity of our method in the form of the autoregressive model.

\section{Regular Sequences Prediction}

Under ideal conditions, the node states can be observed at equal time intervals. In this section, we try to use our model to capture the network dynamics based on regular sequences of node states. In the experiment, we also consider these continuous network dynamics with different underlying networks above. Regarding the experimental setup, we generally follow the settings in NDCN~\cite{zang2020neural}, except for setting a larger time interval for observations, because we consider the sparsity of observations in real life. For each case, we sample 80 snapshots of node states with equal intervals, $\mathbf{x} = \{(t_0, x_0), (t_1, x_1),..., (t_{79}, x_{79}),$ $\{ t_0<...<t_{79}\}$, where time intervals between adjacent observations are equal. The first $80\%$ observations are used for training, while the last $20\%$ are considered as the test set for the regular sequences prediction task. For evaluation, we also employ the \textbf{MAE}(Mean Absolute Error) and \textbf{Normalized L1 loss} metrics to validate the performance of methods.

\textbf{Experimental configuration}. The basic experimental settings are set as the same as that in the previous Section. Consider the time interval between observations is equal. Beyond the baseline methods mentioned above, we also choose temporal-GNN algorithms as the baseline, which are also considered in ~\cite{zang2020neural}. Temporal-GNN algorithms are the combinations of GCN and RNN blocks, and they can not be applied to the interpolation reconstruction task. Only the extrapolation prediction task is considered here. The hidden size of the GCN and RNN blocks are set as $10$ and $5$, respectively. The default parameters are used for NDCN, and the parameters of our model and other GNN-ODE based methods remain the same as in Section $4$.

\subsection{Baseline}

The temporal-GNN algorithms are composed of different RNN cells and GCN blocks. The GCN block simulates and learns network dynamics, while the RNN cell models the sequential connections of node states. Here we just consider GRU, LSTM, and RNN cells. The objective function of the temporal-GNN algorithms is the same as that of NDCN. The details are listed as follows:

\begin{itemize}
\item {\textbf{GRU-GNN}(GRU-G)}. This temporal-GNN is consist of GCN block and GRUCell:
\begin{equation}
\begin{split}
z_t = \operatorname{ReLU}(\Phi x_t W_{e}+b_{e}),\\
h_t = GRUCell(z_t,h_{t-1}),\\
x^{\prime}_{t+1} = h_t W_{d}+b_{d}.
\end{split}
\end{equation}

\item {\textbf{LSTM-GNN}(LSTM-G)}. This temporal-GNN is composed of GCN block and LSTMCell:
\begin{equation}
\begin{split}
z_t = \operatorname{ReLU}(\Phi x_t W_{e}+b_{e}),\\
h_t = LSTMCell(z_t,h_{t-1}),\\
x^{\prime}_{t+1} = h_t W_{d}+b_{d}.
\end{split}
\end{equation}

\item {\textbf{RNN-GNN}(RNN-G)}. This temporal-GNN is consist of GCN block and RNNCell:
\begin{equation}
\begin{split}
z_t = \operatorname{ReLU}(\Phi x_t W_{e}+b_{e}),\\
h_t = RNNCell(z_t,h_{t-1}),\\
x^{\prime}_{t+1} = h_t W_{d}+b_{d}.
\end{split}
\end{equation}

\end{itemize}

\begin{table*}[!t]
\footnotesize
\caption{MAE results($\times 10^{-2}$) in the regular sequences prediction of network dynamics.}
\label{equal1}
\centering
\begin{tabular}{cccccccccc}
\hline
                                                                                         &             & \multicolumn{8}{c}{MAE}                                                                    \\ \hline
                                                                                         &             & AGOG           & NDCN           & CGNN    & WCGNN  & GRAND  & LSTM-GNN & GRU-GNN & RNN-GNN \\ \hline
\multirow{5}{*}{\begin{tabular}[c]{@{}c@{}}Gene   \\      Regulation\end{tabular}}       & Community   & 127.11         & \textbf{90.4}  & 1032.35 & 893.49 & 606.87 & 1773.02  & 1406.7  & 1469.39 \\
                                                                                         & Grid        & \textbf{33.06} & 106.98         & 227.58  & 216.4  & 270.63 & 187.3    & 181.38  & 187.03  \\
                                                                                         & Power Law   & 42.37          & \textbf{33.13} & 453.64  & 286.78 & 440.6  & 310.13   & 335.62  & 322.85  \\
                                                                                         & Random      & 175.41         & \textbf{85.32} & 1063.32 & 961.52 & 689.47 & 2768.31  & 2088.75 & 2219.91 \\
                                                                                         & Small World & \textbf{24.49} & 35.87          & 214.8   & 134.03 & 115.09 & 75.11    & 70.14   & 87.24   \\ \hline
\multirow{5}{*}{\begin{tabular}[c]{@{}c@{}}Kuramoto \\      Model\end{tabular}}          & Community   & \textbf{22.86} & 239.33         & 210.13  & 136.8  & 217.47 & 141.08   & 141.1   & 141.37  \\
                                                                                         & Grid        & \textbf{25.14} & 252.09         & 335.53  & 169.64 & 210.84 & 146.38   & 134.32  & 140.95  \\
                                                                                         & Power Law   & \textbf{24.32} & 201.19         & 191.53  & 122.71 & 244.15 & 138.6    & 136.13  & 142.79  \\
                                                                                         & Random      & \textbf{23.09} & 258.95         & 176.92  & 134.99 & 213.76 & 145.94   & 142.93  & 144.89  \\
                                                                                         & Small World & \textbf{24.03} & 276.11         & 293.77  & 156.23 & 209.02 & 143.97   & 134.7   & 142.84  \\ \hline
\multirow{5}{*}{\begin{tabular}[c]{@{}c@{}}Mutualistic \\      Interaction\end{tabular}} & Community   & \textbf{26.8}  & 45.8           & 348.58  & 373.87 & 168.56 & 177.17   & 41.23   & 94.67   \\
                                                                                         & Grid        & \textbf{39.78} & 112.34         & 246.89  & 206.6  & 85.98  & 196.6    & 205.37  & 216.79  \\
                                                                                         & Power Law   & \textbf{21.4}  & 81.73          & 331.29  & 317.76 & 150.37 & 179.97   & 84.77   & 110.69  \\
                                                                                         & Random      & \textbf{34.47} & 70.18          & 241.21  & 363.66 & 84.28  & 579.91   & 56.17   & 116.97  \\
                                                                                         & Small World & \textbf{47.68} & 54.38          & 160.39  & 144.35 & 92.7   & 167.19   & 153.69  & 167.48  \\ \hline
\end{tabular}
\end{table*}

\begin{table*}[!t]
\footnotesize
\caption{Normalized L1 Loss results($\times 10^{-2}$) in the regular sequences prediction of network dynamics.}
\label{equal2}
\centering
\begin{tabular}{cccccccccc}
\hline
                                                                                         &             & \multicolumn{8}{c}{Normalized L1}                                                     \\ \hline
                                                                                         &             & AGOG           & NDCN          & CGNN  & WCGNN & GRAND & LSTM-GNN & GRU-GNN & RNN-GNN \\ \hline
\multirow{5}{*}{\begin{tabular}[c]{@{}c@{}}Gene   \\      Regulation\end{tabular}}       & Community   & 4.07           & \textbf{2.9}  & 33.09 & 28.64 & 19.45 & 56.86    & 45.07   & 47.11   \\
                                                                                         & Grid        & \textbf{4.86}  & 15.74         & 33.48 & 31.83 & 39.81 & 27.55    & 26.68   & 27.51   \\
                                                                                         & Power Law   & 4.4            & \textbf{3.44} & 47.14 & 29.8  & 45.79 & 32.23    & 34.88   & 33.55   \\
                                                                                         & Random      & 4.46           & \textbf{2.17} & 27.05 & 24.48 & 17.54 & 70.47    & 53.14   & 56.48   \\
                                                                                         & Small World & \textbf{4.34}  & 6.35          & 38.01 & 23.73 & 20.38 & 15.3    & 12.41   & 15.42   \\ \hline
\multirow{5}{*}{\begin{tabular}[c]{@{}c@{}}Kuramoto \\      Model\end{tabular}}          & Community   & \textbf{3.82}  & 40            & 35.13 & 22.87 & 36.34 & 23.55    & 23.55   & 23.59   \\
                                                                                         & Grid        & \textbf{4.11}  & 41.27         & 54.93 & 27.77 & 34.51 & 23.94    & 21.99   & 23.07   \\
                                                                                         & Power Law   & \textbf{4}     & 33.12         & 31.53 & 20.2  & 40.21 & 22.79    & 22.39   & 23.48   \\
                                                                                         & Random      & \textbf{3.81}  & 42.82         & 29.25 & 22.32 & 35.29 & 24.08    & 23.6    & 23.91   \\
                                                                                         & Small World & \textbf{3.93}  & 45.23         & 48.11 & 25.59 & 34.23 & 23.56    & 22.06   & 23.37   \\ \hline
\multirow{5}{*}{\begin{tabular}[c]{@{}c@{}}Mutualistic \\      Interaction\end{tabular}} & Community   & \textbf{1.93}  & 3.3           & 25.1  & 26.91 & 12.13 & 12.78     & 2.97    & 6.82    \\
                                                                                         & Grid        & \textbf{10.4}  & 29.37         & 64.54 & 54.01 & 22.48 & 51.39    & 53.69   & 56.67   \\
                                                                                         & Power Law   & \textbf{2.43}  & 9.27          & 37.56 & 36.02 & 17.05 & 20.4  & 9.61    & 12.55   \\
                                                                                         & Random      & \textbf{2.24}  & 4.55          & 15.65 & 23.61 & 5.47  & 37.64     & 3.63    & 7.59    \\
                                                                                         & Small World & \textbf{15.25} & 17.42         & 51.37 & 46.19 & 29.69 & 53.51    & 49.17   & 53.61   \\ \hline
\end{tabular}
\end{table*}

\subsection{Result}

The results are averaged over 20 independent realizations and shown in Tabs. \ref{equal1} and \ref{equal2}. The best results in each case are denoted in bold. The results show that these two metrics come to the same conclusion. In the Gene Regulation network dynamics, NDCN performs the best in the community network, grid network, and random network, while our model is slightly inferior to NDCN but still comparable. In other scenes of network dynamics, our method demonstrates extraordinary competitiveness and has unparalleled dominance and the best performance. The performance of these temporal-GNN algorithms is unstable and not very good in these circumstances. Besides, one of the key flaws of these temporal-GNN algorithms is that there are too many trainable parameters. WCGNN performs relatively well in Kuramoto model network dynamics than NDCN, while the performance of GRAND is just fine in Mutualistic interaction network dynamics. The results of these methods in other scenarios are very poor. Overall, the performance of other algorithms is uneven and unstable. In the Kuramoto model dynamics, the Normalized L1 loss metric of our model can achieve $16\%$ of that of the second best, GRU-GNN. In the Mutualistic interaction network dynamics, our model can achieve $51.4\%$ of the second best, NDCN, in the MAE metric. To sum up, the results show that our model can have the best or comparable performance and good robustness in almost all cases. Our model can still learn and capture the network dynamics very well in the regular sequences prediction task.

\section{Conclusion}

Revealing the continuous dynamics on complex networks is essential for understanding complex systems. But this task is very hard due to the relatively unknown and partially known equations and high dimensions of complex systems. In addition, the observations of node states are generally non-uniform and very sparse in real life, which also poses a great challenge. In this paper, we propose an Autoregressive GNN-ODE GRU Model(AGOG) to capture the continuous network dynamics based on the observed node states in a data-driven manner. The further predictions are based on the observation history. With the predictions achieved, the true observations at the same time as the prior information are used to update the hidden states for the predictions at the next step. We impose a strong constraint to make the observations into the training process and make the ODE system learn the real continuous network dynamics more easily and accurately. We use a GNN module to simulate and fit the network dynamics, and this module has two parts that describe the interactions between nodes and the dependence of the dynamic evolution, respectively. The hidden state of nodes is specified by the ODE system, and we utilize the augmented ODE to map the hidden state into continuous time. The augmented ODE solves the hidden states in the augmented space, which could achieve lower losses and improve generalization and stability. Then the hidden states are updated through observations by GRUCell. To evaluate the validity of our model, we visualize the learned network dynamics by our model and test it in three tasks: interpolation reconstruction, extrapolation prediction, and regular sequences prediction. The results show that our model can consistently outperform other baseline methods or achieve comparable performance in these dynamics with different underlying networks. Our model can accurately learn the continuous network dynamics with a small error. In future work, we consider introducing the attention mechanism into the learning of network dynamics.

\begin{acknowledgments}
This work was supported by the Project of Science and Technology Commission of Shanghai Municipality, China, under Grant 22JC1401401, the National Natural Science Foundation of China under Grant Nos 61873167, and the Strategic Priority Research Program of Chinese Academy of Sciences under Grant No. XDA27000000(in part).
\end{acknowledgments}

\appendix
\section{Appendix}

\begin{figure*}[!t] 
\centering 
\includegraphics[width=0.75\textwidth]{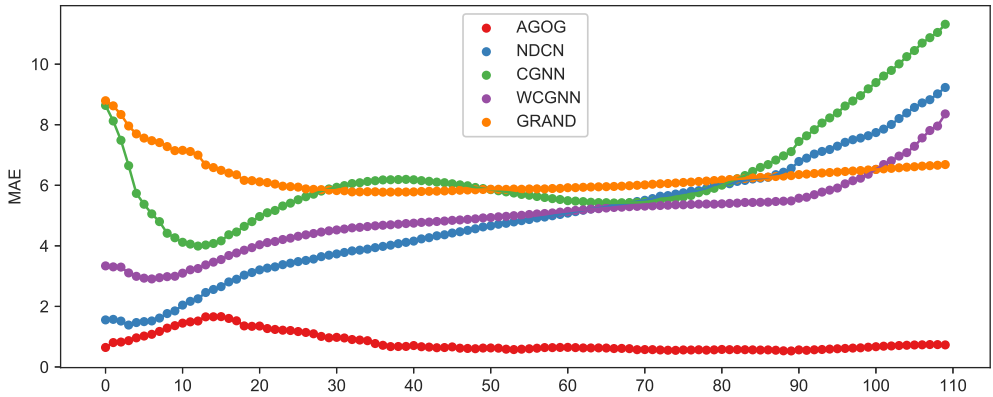} 
\caption{Prediction error for each datapoint of different methods in the Gene regulatory network dynamics with the community network, $P\%$ as $10\%$.} 
\label{Gene_over_time}
\end{figure*}

\begin{figure*}[!t] 
\centering 
\includegraphics[width=0.75\textwidth]{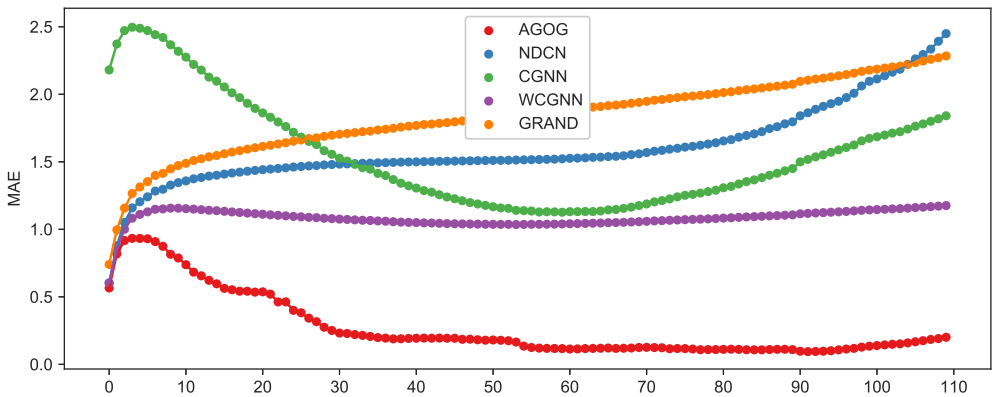} 
\caption{Prediction error for each datapoint of different methods in the Kuramoto network dynamics with the community network, $P\%$ as $10\%$.} 
\label{Kuramoto_over_time}
\end{figure*}

\begin{figure*}[!t] 
\centering 
\includegraphics[width=0.75\textwidth]{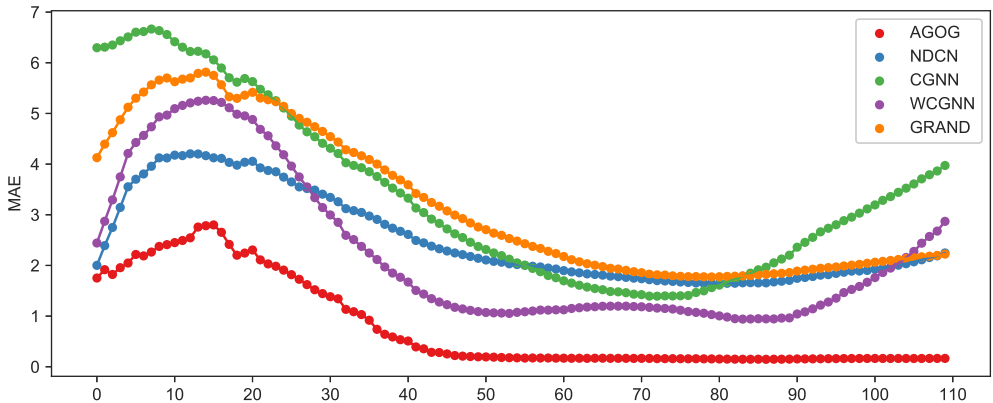} 
\caption{Prediction error for each datapoint of different methods in the Mutualistic interaction network dynamics with the community network, $P\%$ as $10\%$.} 
\label{Mutual_over_time}
\end{figure*}

We show the quantitative results over time of the interpolation reconstruction and extrapolation prediction tasks with $P$ as $10\%$ in the Figs. \ref{Gene_over_time}, \ref{Kuramoto_over_time} and \ref{Mutual_over_time}. Due to space limitations, we just present the results in the community network. In each figure, each point corresponds to the average deviation of the prediction for the node state at a given moment. Here we ignore the specific time of the predicted node states and only consider the temporal order of the predictions. The results over time demonstrate that our method is superior to other methods at any time. The results of interpolation reconstruction and extrapolation prediction tasks with the observed proportion of node states as $30\%$ are shown in Tabs. \ref{inter_0.3}, and \ref{extra_0.3}, respectively. These two metrics all reach the same conclusion. Similar results are found in Tabs. \ref{inter_0.1} and \ref{extra_0.1}. The results show that our model exhibits tremendous power ability and can capture the continuous network dynamics accurately and with a small error.

\begin{table*}[]
\footnotesize
\caption{Results($\times 10^{-2}$) of interpolation reconstruction of network dynamics with the percentage of observed node states as $30\%$.}
\label{inter_0.3}
\centering
\begin{tabular}{cccccccccccc}
\hline
                                                                                       &             & \multicolumn{5}{c}{MAE}                            & \multicolumn{5}{c}{Normalized   L1}           \\ \hline
                                                                                         &             & AGOG           & NDCN   & CGNN   & WCGNN  & GRAND  & AGOG          & NDCN  & CGNN  & WCGNN & GRAND \\ \hline
\multirow{5}{*}{\begin{tabular}[c]{@{}c@{}}Gene   \\      Regulation\end{tabular}}       & Community   & \textbf{44.02} & 167.88 & 565.59 & 466.48 & 598.34 & \textbf{1.81} & 6.89  & 23.13 & 19.06 & 24.44 \\
                                                                                         & Grid        & \textbf{8.08}  & 46.42  & 203.7  & 118.86 & 163.07 & \textbf{1.78} & 10.23 & 45.12 & 26.27 & 36.02 \\
                                                                                         & Power Law   & \textbf{21.31} & 44.84  & 304.94 & 222.21 & 333.06 & \textbf{2.73} & 5.78  & 39.24 & 28.55 & 42.8  \\
                                                                                         & Random      & \textbf{59.58} & 187.32 & 506.02 & 355.29 & 702.61 & \textbf{1.95} & 6.11  & 16.5  & 11.59 & 23.03 \\
                                                                                         & Small World & \textbf{8.27}  & 26.32  & 141.5  & 65.91  & 113.44 & \textbf{1.98} & 6.3   & 33.93 & 15.8  & 27.19 \\ \hline
\multirow{5}{*}{\begin{tabular}[c]{@{}c@{}}Kuramoto \\      Model\end{tabular}}          & Community   & \textbf{16.84} & 142.33 & 150.44 & 107.57 & 194.64 & \textbf{3.5}  & 29.69 & 31.39 & 22.44 & 40.58 \\
                                                                                         & Grid        & \textbf{15.31} & 141.85 & 219.09 & 133.96 & 156.36 & \textbf{3.15} & 29.21 & 45.13 & 27.59 & 32.2  \\
                                                                                         & Power Law   & \textbf{13.33} & 115.04 & 127.48 & 85.31  & 187.11 & \textbf{2.72} & 23.53 & 26.08 & 17.45 & 38.26 \\
                                                                                         & Random      & \textbf{17.96} & 151.52 & 135.19 & 102.82 & 180.39 & \textbf{3.72} & 31.43 & 28.06 & 21.34 & 37.38 \\
                                                                                         & Small World & \textbf{14.4}  & 135.99 & 191.48 & 119.61 & 155.56 & \textbf{2.95} & 27.85 & 39.23 & 24.49 & 31.85 \\ \hline
\multirow{5}{*}{\begin{tabular}[c]{@{}c@{}}Mutualistic \\      Interaction\end{tabular}} & Community   & \textbf{34.71} & 166.05 & 343.73 & 222    & 308.52 & \textbf{3.17} & 15.14 & 31.26 & 20.15 & 28.09 \\
                                                                                         & Grid        & \textbf{9.88}  & 25.63  & 84.11  & 55.69  & 35.29  & \textbf{4.6}  & 11.93 & 39.19 & 25.89 & 16.44 \\
                                                                                         & Power Law   & \textbf{11.55} & 75.35  & 183.26 & 143.99 & 211.69 & \textbf{2.27} & 14.68 & 35.83 & 28.13 & 41.41 \\
                                                                                         & Random      & \textbf{40.64} & 190.7  & 310.74 & 218.81 & 308.77 & \textbf{3.24} & 14.94 & 24.42 & 17.2  & 24.31 \\
                                                                                         & Small World & \textbf{7.08}  & 13.65  & 50.2   & 28.2   & 25.09  & \textbf{4.18} & 8.03  & 29.52 & 16.54 & 14.74 \\ \hline
\end{tabular}
\end{table*}


\begin{table*}[]
\footnotesize
\caption{Results($\times 10^{-2}$) of extrapolation prediction of network dynamics with the percentage of observed node states as $30\%$.}
\label{extra_0.3}
\centering
\begin{tabular}{cccccccccccc}
\hline
                                                                                         &             & \multicolumn{5}{c}{MAE}                            & \multicolumn{5}{c}{Normalized   L1}           \\ \hline
                                                                                         &             & AGOG           & NDCN   & CGNN   & WCGNN  & GRAND  & AGOG          & NDCN  & CGNN  & WCGNN & GRAND \\ \hline
\multirow{5}{*}{\begin{tabular}[c]{@{}c@{}}Gene   \\      Regulation\end{tabular}}       & Community   & \textbf{79.33} & 216.3  & 898.11 & 708.72 & 635.88 & \textbf{2.53} & 6.89  & 28.61 & 22.59 & 20.25 \\
                                                                                         & Grid        & \textbf{9.09}  & 163.71 & 213.6  & 157.98 & 248.28 & \textbf{1.33} & 23.91 & 31.23 & 23.11 & 36.32 \\
                                                                                         & Power Law   & \textbf{28.16} & 50.56  & 431.04 & 282.54 & 435.58 & \textbf{2.92} & 5.25  & 44.75 & 29.33 & 45.22 \\
                                                                                         & Random      & \textbf{81.69} & 221.77 & 919.1  & 640.56 & 615.32 & \textbf{2.07} & 5.62  & 23.3  & 16.24 & 15.61 \\
                                                                                         & Small World & \textbf{10.69} & 36.64  & 192.26 & 89.6   & 107.77 & \textbf{1.89} & 6.49  & 34.05 & 15.89 & 19.1  \\ \hline
\multirow{5}{*}{\begin{tabular}[c]{@{}c@{}}Kuramoto \\      Model\end{tabular}}          & Community   & \textbf{13.7}  & 182.87 & 173.01 & 115.78 & 244.22 & \textbf{2.27} & 30.39 & 28.77 & 19.23 & 40.59 \\
                                                                                         & Grid        & \textbf{14.66} & 176.91 & 304.4  & 157.57 & 206.63 & \textbf{2.42} & 29.22 & 50.25 & 26.01 & 34.1  \\
                                                                                         & Power Law   & \textbf{12.59} & 170.19 & 156.69 & 105.95 & 258.42 & \textbf{2.05} & 27.84 & 25.62 & 17.32 & 42.23 \\
                                                                                         & Random      & \textbf{13.91} & 178.5  & 138.29 & 107.81 & 225.84 & \textbf{2.29} & 29.5  & 22.87 & 17.83 & 37.3  \\
                                                                                         & Small World & \textbf{14.1}  & 181.54 & 250.1  & 141.32 & 208.99 & \textbf{2.31} & 29.76 & 40.98 & 23.15 & 34.23 \\ \hline
\multirow{5}{*}{\begin{tabular}[c]{@{}c@{}}Mutualistic \\      Interaction\end{tabular}} & Community   & \textbf{19.58} & 125.3  & 332.21 & 209.61 & 147.17 & \textbf{1.41} & 9.01  & 23.9  & 15.09 & 10.59 \\
                                                                                         & Grid        & \textbf{25.69} & 170.63 & 246.87 & 211.63 & 78.17  & \textbf{6.64} & 44.04 & 63.57 & 54.49 & 20.13 \\
                                                                                         & Power Law   & \textbf{22.83} & 91.07  & 278.12 & 220.83 & 130.34 & \textbf{2.59} & 10.32 & 31.52 & 25.03 & 14.77 \\
                                                                                         & Random      & \textbf{15.71} & 146.11 & 229.37 & 222.86 & 102.63 & \textbf{1.02} & 9.46  & 14.88 & 14.43 & 6.65  \\
                                                                                         & Small World & \textbf{16.25} & 74.67  & 159.9  & 148.57 & 94.78  & \textbf{5.09} & 23.28 & 50.01 & 46.4  & 29.64 \\ \hline
\end{tabular}
\end{table*}

\bibliography{sample-gnn_ode_abbr}

\end{document}